\documentclass[conference]{IEEEtran}
\pdfoutput=1
\IEEEoverridecommandlockouts
\usepackage{cite}
\usepackage{amsmath,amssymb,amsfonts}
\usepackage[ruled, vlined]{algorithm2e}
\usepackage{graphicx}
\usepackage{textcomp}
\usepackage{xcolor}
\usepackage{url}
\usepackage{booktabs}
\usepackage{subcaption}
\usepackage{multirow}
\usepackage{tabularx}
\usepackage{cleveref}
\def\BibTeX{{\rm B\kern-.05em{\sc i\kern-.025em b}\kern-.08em
    T\kern-.1667em\lower.7ex\hbox{E}\kern-.125emX}}
\begin{document}

\title{Ensemble Deep Learning on Time-Series Representation of Tweets for Rumor Detection in Social Media}

\author{\IEEEauthorblockN{Chandra Mouli Madhav Kotteti, Xishuang Dong, Lijun Qian}
\IEEEauthorblockA{Center of Excellence in Research and Education for Big Military Data Intelligence (CREDIT Center) \\
Prairie View A\&M University,   Texas A\&M University System \\
Prairie View, TX 77446, USA\\
ckotteti@student.pvamu.edu, xidong@pvamu.edu, liqian@pvamu.edu}}

\maketitle

\begin{abstract}
Social media is a popular platform for timely information sharing. One of the important challenges for social media platforms like Twitter is whether to trust news shared on them when there is no systematic news verification process. On the other hand, timely detection of rumors is a non-trivial task, given the fast-paced social media environment. In this work, we proposed an ensemble model, which performs majority-voting on a collection of predictions by deep neural networks using time-series vector representation of Twitter data for timely detection of rumors. By combining the proposed data pre-processing method with the ensemble model, better performance of rumor detection has been demonstrated in the experiments using PHEME dataset. Experimental results show that the classification performance has been improved by $7.9\%$ in terms of micro F1 score compared to the baselines.
\end{abstract}

\begin{IEEEkeywords}
social media, rumor detection, time-series data, machine learning, deep learning, Twitter
\end{IEEEkeywords}

\section{Introduction}
\label{sec:intro}
Over the past few decades social media have emerged out as the primary means for news creation as well as for news consumption. Given the speed at which information travels on social media it is very easy to propagate any type of news and it can be consumed instantly across the globe at the early stages of its propagation process. However, the biggest challenge for news spreading on social media is how to verify whether that news is correct or not. Even though social media outperforms traditional media in many aspects, the key difference between them is that 
the news is verified for its truthfulness before it gets proliferated in traditional media, while it is not the case for 
social media. Thus, any piece of information can be easily spread on social media regardless of its truthfulness.

Furthermore, information shared on social media propagates rapidly and increases the difficulty in verifying its credibility in near real time. A rumor is defined as a ``circulating story of questionable veracity, which seems credible but hard to verify \cite{zubiaga2015towards}, and produces sufficient skepticism and anxiety'', and it could have truth values such as true, false or unverified \cite{shu2017fake}. Detection of rumors in social media has a lot of importance among research communities because unverified information may be easily disseminated over a large network, and rumors may spread misinformation or disinformation\footnote{Misinformation means information that is incorrect in its nature and disinformation means information that is used to deceive its consumers.}, which are forms of false information \cite{qazvinian2011rumor, kshetri2017economics}. 

If the spread of false information is not stopped early it may cause turmoil in the society. In case of time critical events,  the effects may be dreadful. So detecting rumors in social media must be done in a timely fashion. Recently machine learning and deep learning gained huge popularity in addressing rumor detection in social media \cite{reshi2019rumor}, and they typically applies trained classification models to predict new data samples as rumors or non-rumors \cite{wang2019rumor}. One of the  main concerns for applying these techniques is to find a dataset with good quality. On the other hand, performing extensive feature engineering on the dataset to extract a variety of useful features for the rumor identification problem may help in improving a classification model's performance. However, it will significantly slow down the training procedure since employing complex features in training process is cumbersome in terms of computational complexity and availability of hardware resources to deal with extremely large sized feature set \cite{kotteti2018multiple}. Hence, extensive feature engineering may not be suitable for timely rumor detection. 

In this paper, we explore the temporal features of Twitter data for timely detection of rumors in social media. 
Tweet creation timestamp can readily be extracted from tweets, and there is no time delay to collect  timestamp features and no sophisticated data pre-processing is required to convert them into useful features to train a classification model. Based on this observation, we proposed an ensemble based multiple time-series analysis model using deep learning models for timely detection of rumors in social media. Specifically, we generated time-series data by transforming Twitter conversations, where each conversation contains a list of tweets, into times-series vectors that contain reaction counts as features, and fed them as input to deep learning models. The contributions of our proposed method are:
\begin{itemize}
\item With the proposed method, computational complexity can be significantly reduced as we just need timestamps of tweets rather than their contents or user social engagements to perform feature extraction. Moreover, the extracted feature set is of numeric type, which is amicable to classification models.
\item Our proposed ensemble model improves the performances of classification models since it uses the majority-voting scheme on multiple neural networks that are part of the ensemble model and takes advantage of their individual strengths.
\item We validated our proposed method on the PHEME\footnote{\url{https://figshare.com/articles/PHEME_dataset_for_Rumour_Detection_and_eracity_Classification/6392078}} dataset and the performance results demonstrate the effectiveness of the proposed scheme.
\end{itemize}

\section{Problem formulation}
\label{sec:probfor}

\subsection{Rumor detection}
\label{subsec:probdef}
Rumor detection involves identifying whether a data sample is a rumor or not. In machine learning, this kind of problem is termed as a classification task, in which the classification model gets trained with adequate number of training samples and tries to classify a never before seen testing sample as rumor or not. Therefore, the problem is given by  $\hat{y} = f(X)$, where $f$ is the classification model and $X$ is a completely new data sample (a Twitter conversation sample that is transformed into a time-series vector) to it, and $\hat{y}$ is the prediction of the classification model and it has only two values since the PHEME dataset has two classes. In our work, we used $0$'s and $1$'s to represent non-rumor and rumor samples, respectively, i.e., $\hat{y} \in \{0, 1\}$.

\subsection{General features of tweets}
\label{subsec:genfea}
Typically, for a classification task using machine learning or deep learning requires extraction of useful features from the dataset.  A variety of features can be extracted from Twitter data, for example, four types of features are extracted  from Twitter data for the study on spread of anomalous information in social media~\cite{zhao2014fluxflow}: user profile features (users' friends and followers count), user network features (users' EgoNet features), temporal features (retweet count), and content features (e.g. whether a tweet has question mark). However, based on the theories of rumor propagation, authors in \cite{kwon2013prominent} considered temporal features as one of the key properties for studying spread of rumors since according to social psychologists rumormongers have a short attention. In this work, for the fast detection of rumors on social media, we solely focused on the temporal features of Twitter data, which are the creation timestamps of tweets. These timestamps can be readily fetched, and our work strictly relies on them for generation of time-series data, which involves simple calculations i.e. counting of number of tweets for given time interval limits.

\subsection{Feature extraction}
\label{subsec:feaextract}
In general, for Twitter data we use a parser to read and extract required information from it by depending up on its data type. In our work, the Twitter data we utilized is available in \textit{JSON} format and we used a suitable parser to read that information and extracted our required features, which are the creation timestamps of tweets.

\section{Ensemble learning}
\label{sec:ensdl}

\subsection{Overview of ensemble learning}
\label{subsec:enslearn}
Ensemble learning is a concept in which many weak or base learners try to solve a single problem. An ensemble contains a number of base learners and its generalization ability is powerful than that of the base learners \cite{zhou2015ensemble}. Ensemble methods work on a set of hypotheses derived from training data rather than relying on one hypothesis. Constructing ensembles is a two-step process. At first, required number of base learners are produced. Secondly, all the base learners are grouped and typically majority voting is applied for classification problems, and weighted averaging combination schemes are used for regression problems. Popular ensemble methods are boosting \cite{freund1997decision}, bagging \cite{breiman1996bagging}, and stacking \cite{wolpert1992stacked}. Boosting method focuses on fitting multiple weak learners sequentially, where each model in a sequence gives more emphasis to the data samples that were badly treated by its previous model. AdaBoost \cite{freund1997decision} algorithm is a good example of boosting, which is simple and can be applied to data that is numeric, textual, etc. In bagging method, multiple bootstrap samples are generated from the training data, and an independent weak learner is fitted for each of these samples. Finally, all the predictions of weak learners are aggregated to determine the most-voted class. RandomForests \cite{breiman2001random} algorithm is good example of bagging method, which is one of the most accurate learning algorithms and runs efficiently on large databases. In stacking method, by using different learning algorithms, multiple first-level individual learners are created, and these learners are grouped by a second-level learner (meta-learner) to output a prediction \cite{wolpert1992stacked}.

\subsection{Bagging learning}
\label{subsec:baglearn}
Bagging learning has been studied extensively in the literature. Bagging also known as bootstrap aggregation is a popular ensemble method that is useful in reducing the high variance of machine learning algorithms. In bagging technique, several datasets are derived from the original training data set by employing sampling with replacement strategy that means some observations in the derived datasets may be repeated. These datasets are used to train classification or regression models, and outputs of them are typically weighted averaged for regression cases or majority voted for classification problems.

Majority voting grouping technique is used in \cite{gaikwad2015intrusion, tuysuzoglu2016ensemble}. In \cite{gaikwad2015intrusion}, bagging method of ensemble is used with REPTree as base classifier for intrusion detection system, and compared to other traditional machine learning techniques. It is shown that ensemble bagging method achieved high classification accuracy by employing NSL\_KDD dataset. Authors in \cite{tuysuzoglu2016ensemble}, proposed to use dictionary learning with random subspace and bagging methods, and introduced Random Subspace Dictionary Learning (RDL) and Bagging Dictionary Learning (BDL) algorithms. Their experimental analysis concluded that ensemble based dictionary learning methods performed better than that of single dictionary learning.

Weighted averaging grouping technique is employed in \cite{li2010accurate, linghu2010constructing}. In \cite{li2010accurate}, Neural Network Ensemble (NNE) approach is proposed to improve generalization ability of neural networks, and to reduce the calculation errors of Density Functional Theory (DFT). It is shown that both simple averaging and weighted averaging grouping techniques helped in improving DFT calculation results. Authors in \cite{linghu2010constructing}, proposed a method for improving image classification performance using SVM ensembles. Optimal weights for the base classifiers in the SVM ensemble are estimated by solving a quadratic programming problem. These weights are then used to combine the base classifiers to form an SVM ensemble.


Optimization of a generic bagging algorithm is studied in \cite{zeng2010optimization}. Authors added an optimization process into the bagging algorithm that focuses on selecting better classifiers, which are relatively efficient, and proposed a Selecting Base Classifiers on Bagging (SBCB) algorithm. Experimental results proved that their SBCB algorithm performed well than generic bagging approach.

\subsection{Deep bagging learning}
\label{subsec:deepbaglearn}
Because deep neural networks are nonlinear methods and have high variance,  ensemble learning can 
combine the predictions of multiple neural network models in order to achieve less variance among the predictions and to decrease the generalization error. Ensemble method is applied to neural networks mainly by (1) varying training data (data samples used to train models in the ensemble are varied), (2) varying choice of the models in the ensemble, and (3) varying the combination techniques that determine how outputs of ensemble members are combined.

In \cite{chen2019deep}, authors proposed a method that uses Convolutional Neural Network (CNN) and deep residual network (ResNET) ensemble-based classification methods for Hyperspectral Image (HSI) classification. Their proposed method uses deep learning techniques, random feature selection, and majority voting strategy. Moreover, a transferring deep learning ensemble is also proposed to make use of the learned weights of CNNs. In \cite{islam2008bagging}, two cooperative algorithms namely NegBagg (bagging is used) and NegBoost (boosting is used) are proposed for designing neural network (NN) ensembles. These algorithms use negative correlation algorithm while training NNs in the ensemble. Applying these models to well-known problems in machine learning showed that with lesser number of training epochs compact NN ensembles with good generalization are produced.

In \cite{shi2009bagging}, bagging ensemble is proposed to improve the prediction performance of artificial neural networks (ANN) to tackle bankruptcy prediction problem. Experimental results showed that proposed method improved performance of ANNs. Bagging technique using an ANN is proposed to address imbalance datasets on clinical prediction in \cite{fakhruzi2018artificial}, and experimental results showed that this method improved the prediction performance.

\subsection{Overview of the proposed model}
\label{subsec:propmodel}
Our proposed model has two key components: data pre-processing method and ensemble model. Firstly, raw Twitter conversations are processed to transform them into required data format and then the transformed data is supplied to the ensemble model to perform the classification task. The ensemble model consists of six different neural networks (base learners) that are trained using the generated time-series data and their predictions are grouped such that majority voting scheme is applied on them to determine the outcome as rumor or non-rumor.

\section{Methodology}
\label{sec:method}
\begin{figure*}[hbt!]
\centering
\includegraphics[width=\textwidth, height=7cm]{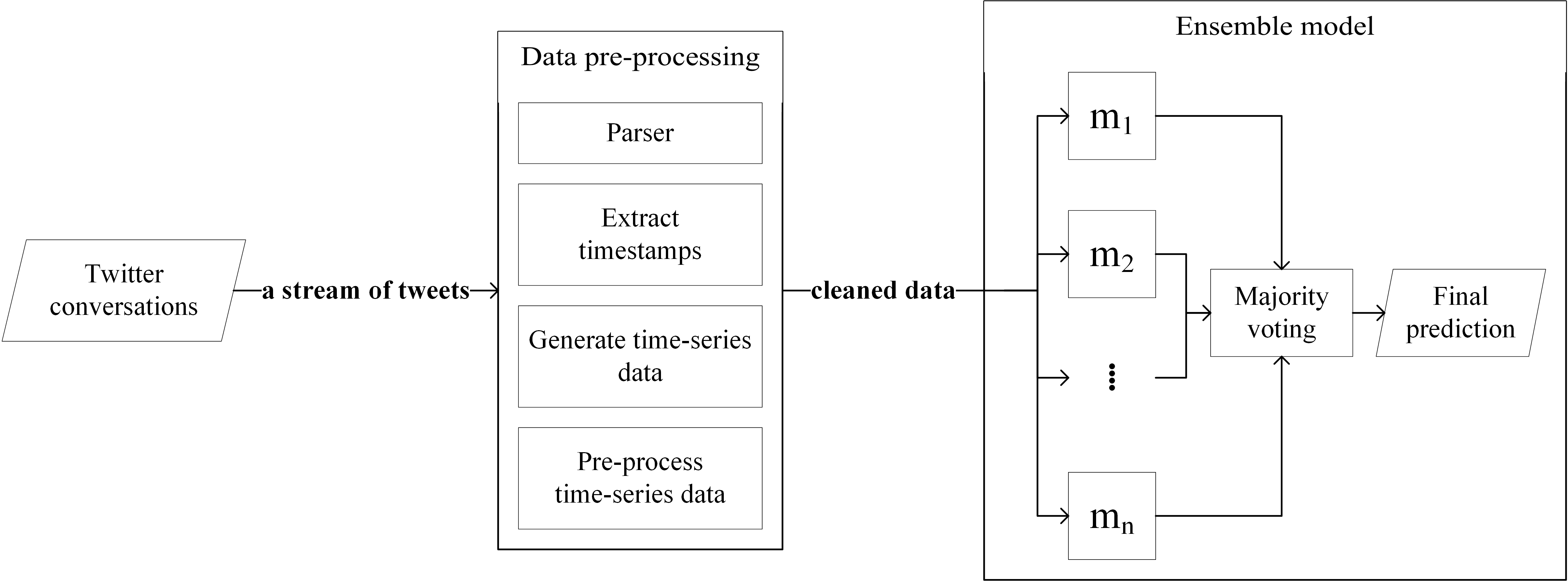}
\caption{Proposed model for rumor classification taking Twitter conversations as input, which are cleaned in the data pre-processing block and fed as input to the ensemble model that performs the majority voting to determine the final prediction}
\label{fig:propmodel}
\end{figure*}
The structure of our proposed model is shown in Fig. \ref{fig:propmodel}. The model takes Twitter conversations as input, where each conversation is a stream of tweets that contains source-tweet and its corresponding reactions. In data pre-processing stage, we parse every tweet and extract its creation timestamp value. Once all tweets are parsed, we generate time-series data for different time intervals and conduct data cleaning on it. Then we fed that cleaned data as input to the ensemble model. The ensemble model has  $n$ base learners, which are $n$ different neural networks that are represented as $m_{1}, m_{2}, \cdots, m_{n}$, where each of them yields its individual prediction results (i.e. $r_{1}, r_{2}, \cdots, r_{n}$). Finally, we perform the majority-voting process on all the predictions of those base learners, i.e., summing up all the prediction results and deciding the final prediction result as $0$ (non-rumor) if total sum is less than $\lfloor n/2 \rfloor +1$ or as $1$ (rumor) otherwise.

\subsection{Neural networks models considered}
The ensemble model constitutes base learners designed using Recurrent Neural Network (RNN), Long Short-Term Memory (LSTM), Gated Recurrent Unit (GRU), and Bi-directional Recurrent Neural Network (Bi-RNN).
Six base learners are designed in this work: BiGRU, BiLSTM, GRU, LSTM, LG (a combination of LSTM and GRU layers), and RNN.

\subsubsection{RNN}
An RNN is a type of neural network that processes sequences by iterating through the sequence elements \cite{book:nns}. Typically, it consists of a hidden state $\textbf{h}$, and an optional output $\textbf{y}$ for a given variable length input sequence $\textbf{x} = (x_{1}, \cdots, x_{T})$. At each time $t$, the hidden state $\textbf{h}_{(t)}$ is given by \cite{cho2014learning}:
\begin{equation}
\textbf{h}_{(t)} = f(\textbf{h}_{(t-1)}, x_{t}),
\label{eq:rnn}
\end{equation}
where $f$ is a non-linear activation function. We used \textit{Keras'} \texttt{SimpleRNN} \cite{keras} layer in our experiments.

\subsubsection{LSTM}
It is a special type of RNN and has been developed by Hochreiter and Schmidhuber in 1997 \cite{hochreiter1997long}. It consists of four major components, which are called as cell, forget gate, input and output gates. Component cell functions to memorize values over arbitrary time intervals and three gates regulate flow of information into or our of the cell \cite{book:nns}. Each $j^{th}$ LSTM unit has a memory $c^{j}_{t}$ at time $t$ and the output $h^{j}_{t}$ is given by \cite{chung2014empirical}:
\begin{equation}
h^{j}_{t} = o^{j}_{t} \tanh(c^{j}_{t}),
\label{eq:lstm}
\end{equation}
where $o^{j}_{t}$ is an output gate.

\subsubsection{GRU}
Chung et al. in 2014 \cite{chung2014empirical} developed Gated Recurrent Unit, which has architecture similar to LSTM. There is no output gate in GRU, which means it has lesser number of parameters than LSTM. To control flow of information it uses update and reset gates, these gates decide how much of past information should be passed along to future or discarded \cite{book:nns}. Linear interpolation between $h^{j}_{t-1}$ and $\tilde{h}^{j}_{t}$, which are previous activation and candidate activation respectively at time $t$ is the activation $h^{j}_{t}$ \cite{chung2014empirical}:
\begin{equation}
h^{j}_{t} = (1-z^{j}_{t})h^{j}_{t-1}+z^{j}_{t}\tilde{h}^{j}_{t},
\end{equation}
where $z^{j}_{t}$ is an update gate.

\subsubsection{Bi-RNN}
A traditional RNN processes the time-steps in order, whereas Bi-RNN \cite{schuster1997bidirectional} exploits the order sensitivity present in RNN and the input sequence can be processed in forward and reverse directions. It may have overfitting issues as it has twice the number of parameters of a traditional RNN, however, overfitting problem can be controlled by employing good regularization techniques \cite{book:nns}. We employed RNN variants GRU and LSTM layers in our experiments. The forward and backward hidden sequences (i.e. $\stackrel{\rightarrow}{h}$ and $\stackrel{\leftarrow}{h}$) for Bi-RNNs are given by:
\begin{equation}
\stackrel{\rightarrow}{h}_{t} = \mathcal{H}(W_{x\stackrel{\rightarrow}{h}}x_{t}+W_{\stackrel{\rightarrow}{h}\stackrel{\rightarrow}{h}}\stackrel{\rightarrow}{h}_{t-1}+b_{\stackrel{\rightarrow}{h}})
\end{equation}
\begin{equation}
\stackrel{\leftarrow}{h}_{t} = \mathcal{H}(W_{x\stackrel{\leftarrow}{h}}x_{t}+W_{\stackrel{\leftarrow}{h}\stackrel{\leftarrow}{h}}\stackrel{\leftarrow}{h}_{t+1}+b_{\stackrel{\leftarrow}{h}}),
\end{equation}
where the $W$ terms denote weight matrices, the $b$ terms denote bias vectors, and $\mathcal{H}$ is the hidden layer function \cite{graves2013hybrid}.

Once the base learners ($m_{1}, m_{2}, \cdots, m_{n}$) complete their training procedures, the ensemble model combines all of their predictions, and performs majority voting procedure on them to determine the ensemble model's evaluation metrics. At first, we created our proposed ensemble model that consists of six base learners. Then we experimented on the proposed model by tuning its hyperparameters such as its batch input size and learning rate, and also created new ensemble models using RNN, LSTM, and GRU layers to obtain a comprehensive set of results to analyze and determine the effectiveness of each ensemble model in efficiently detecting rumor Twitter conversations. Variants of the ensemble model will also have six base learners.

\subsection{Implementation-1}
In implementation 1, each of five base learners (BiGRU\_1, BiLSTM\_1, GRU\_1, LSTM\_1, and simple RNN\_1) has one hidden layer and the sixth based learner (LG\_1) has two hidden layers, followed by one output Dense layer. For all the base learners, the number of hidden layer units is determined based on the integer value obtained from $(seq\_len+2)/2$, where $seq\_len$ is the length of the feature set (i.e., vector length of the time-series data) and constant $2$ is used because number of classification outputs are two (rumor and non-rumor). We considered this approach by following one of rule-of-thumb methods, which states that the number of hidden layer neurons should be between the size of the input layer and the size of the output layer \cite{heaton2008introduction}. $RandUniform$ kernel initializer is used for all the hidden layers with values $(-0.5, 0.5)$. $sigmoid$ activation is applied only to RNN model's hidden layer, and Flatten layer is applied only to BiGRU and BiLSTM models to flatten the data before the final output Dense layer that is activated using $softmax$ function. Adam optimizer is used with learning rate $1.00E-05$ along with \textit{categorical cross-entropy} loss function. Batch input size is set to $32$ and number of epochs is $300$. We did not use Dropout technique with these models since their architectures are simple, and using it may cause under-fitting issues. The variants of the proposed model follow the same neural network design except for the hyperparameter that is tuned, for example, batch input size and learning rate.

\begin{table}[htbp]
  \centering
  \caption{Configurations of NN models}
    \begin{tabular}{c|c|c|c}
    \toprule
    \textbf{NN model} & \textbf{\# of hidden layers} & \textbf{Hidden layer units} & \textbf{Dropout} \\
    \midrule
    RNN\_1   & \multirow{3}[6]{*}{1} & \multirow{3}[6]{*}{$(seq\_len+2)/2$} & \multirow{3}[6]{*}{N/A} \\
\cmidrule{1-1}    GRU\_1   &       &       &  \\
\cmidrule{1-1}    LSTM\_1  &       &       &  \\
    \midrule
    RNN\_2 & \multirow{3}[6]{*}{3} & \multirow{3}[6]{*}{16, 32, 64} & \multirow{6}[12]{*}{0.25} \\
\cmidrule{1-1}    GRU\_2 &       &       &  \\
\cmidrule{1-1}    LSTM\_2 &       &       &  \\
\cmidrule{1-3}    RNN\_3 & \multirow{3}[6]{*}{2} & \multirow{3}[6]{*}{64, 32} &  \\
\cmidrule{1-1}    GRU\_3 &       &       &  \\
\cmidrule{1-1}    LSTM\_3 &       &       &  \\
    \bottomrule
    \end{tabular}%
  \label{tab:confignn}%
\end{table}%

\subsection{Implementation-2}
Six base learners (RNN\_1, RNN\_2, RNN\_3, GRU\_1, GRU\_2, and GRU\_3) have been used in this implementation.
To create new ensembles with new base learners , we used RNN, LSTM, and GRU layers. For instance, for base learners designed using RNN layer, we reused the RNN\_1 base learner designed for implementation 1, and created new base learners by adding extra hidden layers with increasing (RNN\_2) and decreasing (RNN\_3) number of hidden layer units. The configurations of the base learners are shown in TABLE \ref{tab:confignn}. All these base learners are having final output dense layer with $softmax$ activation and loss function as \textit{categorical cross-entropy}. $RandUniform$ kernel initializer with values $(-0.5, 0.5)$. Number of training epochs is set to $300$. For RNN\_1, GRU\_1, and LSTM\_1 base learners in TABLE \ref{tab:confignn}, $seq\_len$ is the length of the feature set. 

\subsection{Implementation-3}
Similar to implementation 2, six base learners (RNN\_1, RNN\_2, RNN\_3, LSTM\_1, LSTM\_2, and LSTM\_3) are employed in implementation 3. The hyperparameters have been set similarly.

\section{Dataset}
\label{sec:dataset}

\subsection{PHEME dataset}
\label{sec:pheme}
\begin{figure}[hbt!]
\centering
\includegraphics[width=0.47\textwidth, keepaspectratio]{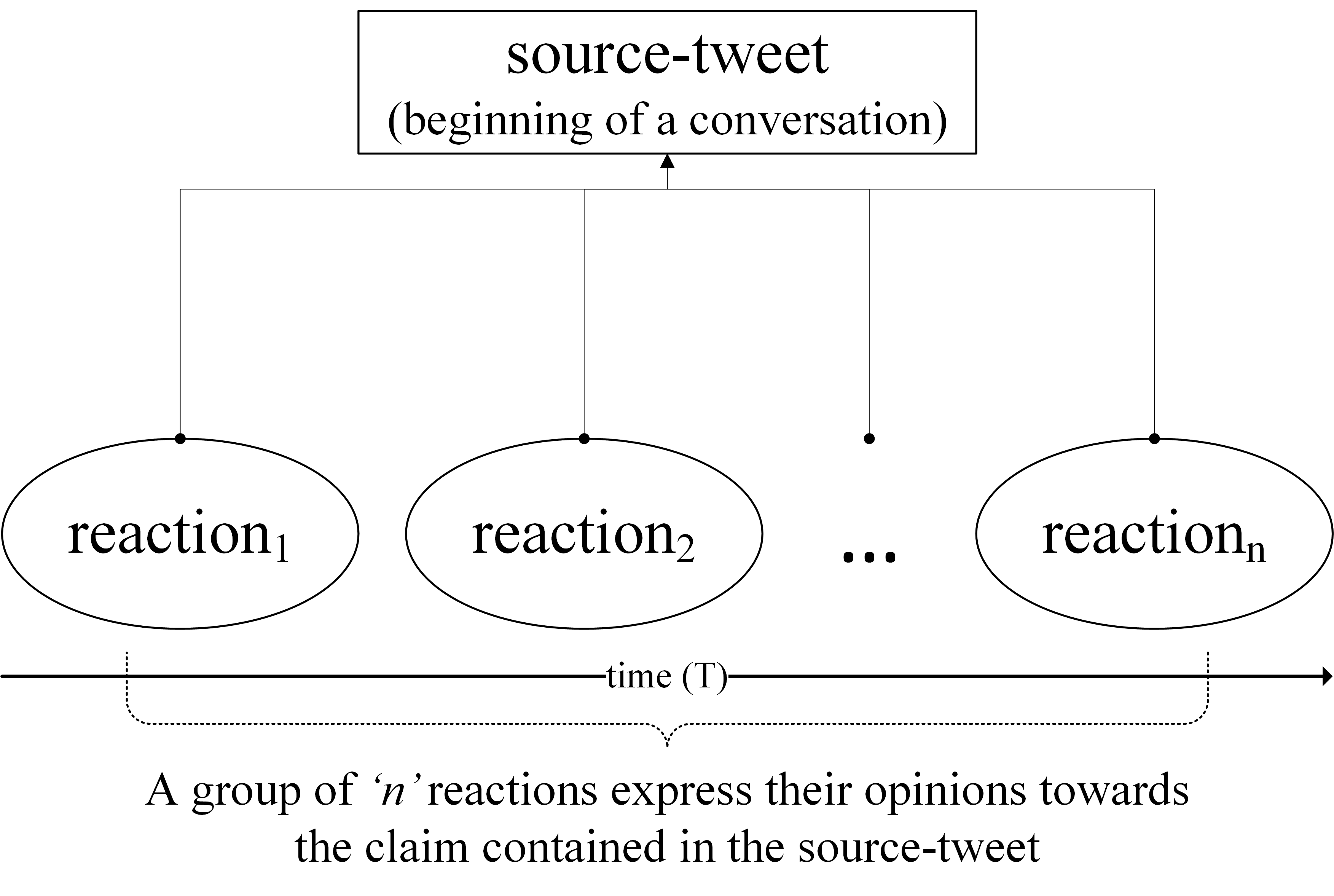}
\caption{Structure of a Twitter conversation sample}
\label{fig:structure}
\end{figure}

In this work, we used the PHEME \cite{kochkina2018all} dataset of rumors and non-rumors, which consists of Twitter conversations for nine different newsworthy events. The distribution of the dataset is shown in TABLE \ref{tab:9events}. The basic structure of conversation samples is shown in Fig. \ref{fig:structure}. Each conversation sample has a source-tweet and a set of reactions along time, where reactions express their opinions towards the claim contained in the source-tweet.

\begin{table}[htbp]
  \centering
  \caption{The PHEME dataset with nine events}
    \begin{tabular}{cccc}
    \toprule
    \textbf{Event} & \textbf{Rumors} & \textbf{Non-rumors} & \textbf{Total} \\
    \midrule
    Charlie Hebdo & 458   & 1,621 & 2,079 \\
    \midrule
    Ferguson & 284   & 859   & 1,143 \\
    \midrule
    Germanwings-crash & 238   & 231   & 469 \\
    \midrule
    Ottawa shooting & 470   & 420   & 890 \\
    \midrule
    Sydney siege & 522   & 699   & 1,221 \\
    \midrule
    Gurlitt & 61    & 77    & 138 \\
    \midrule
    Putin missing & 126   & 112   & 238 \\
    \midrule
    Prince Toronto & 229   & 4     & 233 \\
    \midrule
    Ebola Essien & 14    & 0     & 14 \\
    \midrule
    \textbf{Total} & 2,402 & 4,023 & 6,425 \\
    \bottomrule
    \end{tabular}%
  \label{tab:9events}%
\end{table}%

As shown in TABLE \ref{tab:9events}, this dataset exhibits severe event-wise and class-wise unbalanced nature. For example, event Charlie Hebdo is dominant over all other events present in the dataset in terms of number of samples causing event-wise unbalance. In general, the number of non-rumor class samples are way more than the number of rumor class samples, which is class-wise unbalance in the dataset.

\begin{table}[htbp]
  \centering
  \caption{Distribution of the PHEME dataset with seven events}
    \begin{tabular}{cccc}
    \toprule
    \textbf{Event} & \textbf{Rumors} & \textbf{Non-rumors} & \textbf{Total} \\
    \midrule
    Charlie Hebdo & 458 (22.03\%) & 1,621 (77.97\%) & 2,079 \\
    \midrule
    Ferguson & 284 (24.85\%) & 859 (75.15\%) & 1,143 \\
    \midrule
    Germanwings Crash & 238 (50.75\%) & 231 (49.25\%) & 469 \\
    \midrule
    Gurlitt & 61 (44.20\%) & 77 (55.80\%) & 138 \\
    \midrule
    Ottawa Shooting & 470 (52.81\%) & 420 (47.19\%) & 890 \\
    \midrule
    Putin missing & 126 (52.94\%) & 112 (47.06\%) & 238 \\
    \midrule
    Sydney Siege & 522 (42.75\%) & 699 (57.25\%) & 1,221 \\
    \midrule
    \textbf{Total} & 2,159 (34.95\%) & 4,019 (65.05\%) & 6,178 \\
    \bottomrule
    \end{tabular}%
  \label{tab:7events}%
\end{table}%

In our analysis, we removed events Prince Toronto and Ebola Essien as they have extremely unbalanced proportions of rumors and non-rumors, and trimmed down the dataset to seven events. For example, Ebola Essien event has zero number of non-rumor class samples. The basic statistics of the PHEME dataset with seven events are shown in TABLE \ref{tab:7events}. Overall, the PHEME seven events dataset has $6,178$ data samples, in which non-rumor class samples are almost double the number of rumor class samples.

\subsection{Generation of time-series data}
\label{subsec:gentimedata}
In this paper, we explore the temporal features of Twitter data for timely detection of rumors in social media. 
Specifically, we generated time-series data by transforming Twitter conversations, where each conversation contains a list of tweets, into times-series vectors that contain reaction counts as features, and fed them as input to deep learning models.
We transformed each of the Twitter conversation sample present in the PHEME seven events dataset into time-series vector for each time interval $T$, where $T = \{2, 5, 10, 30, 60\}$ minutes. After successful transformation of all conversations into time-series data, each vector represents one whole conversation and each of its values are the total reaction counts with respect to $T$. 

Denote $ E = \{ e_{i} \} $ the set that contains data of seven events present in the dataset, then for each event data $e_{i} $, $ c_{ij}$ is a conversation sample related to that event. As the dataset has conversations separated by event, we iterated over all the events one-by-one. In each iteration, for every conversation sample present in them, we extracted timestamp of its source-tweet $ timeSource $ (starting point of the conversation) and its $ timeReactions = \{tr_{1}, tr_{2}, \cdots, tr_{n}\}$, which is a set of timestamps of all the reactions corresponding to that source-tweet. For a conversation sample, its length $ N\left(c\right) $ is determined by,
\begin{equation}
N\left(c\right) = \big \lceil \frac{\max\left(timeReactions\right) - timeSource}{T} \big \rceil
\end{equation}
Assume $ c $ represents a conversation sample, if $ \left(a, b\right] $ is the time interval limit for $ k\mbox{-}th $ interval, where $ k = 1, 2, \cdots, N\left(c\right) $ then the total reactions count for that time interval is given by,
\begin{equation}
count_{k} = \mathbf{card}(Q)
\end{equation}
where $ Q \subset timeReactions $ and $ Q = \big \{ x \quad | \quad x > a \land x \leq b \big \} $, here $ x $ is the timestamp of a reaction (tweet) and cardinality is the measure of the size of set $ Q $, and the transformed vector representation is as follows:
\begin{equation}
V\left(c\right) = \left[count_{k} \quad count_{k+1} \quad \cdots \quad count_{N}\right]
\end{equation}
The final vector representation of all conversation samples for each event is given by,
\begin{align}
\label{eq:finalvect}
e_{i} &= \begin{bmatrix}
			V(c_{1}) \\
			V(c_{2}) \\
			\vdots \\
			V(c_{n})\\
		\end{bmatrix}
\end{align}
\begin{figure}[hbt!]
\centering
\includegraphics[width=0.2\textwidth, height=13cm]{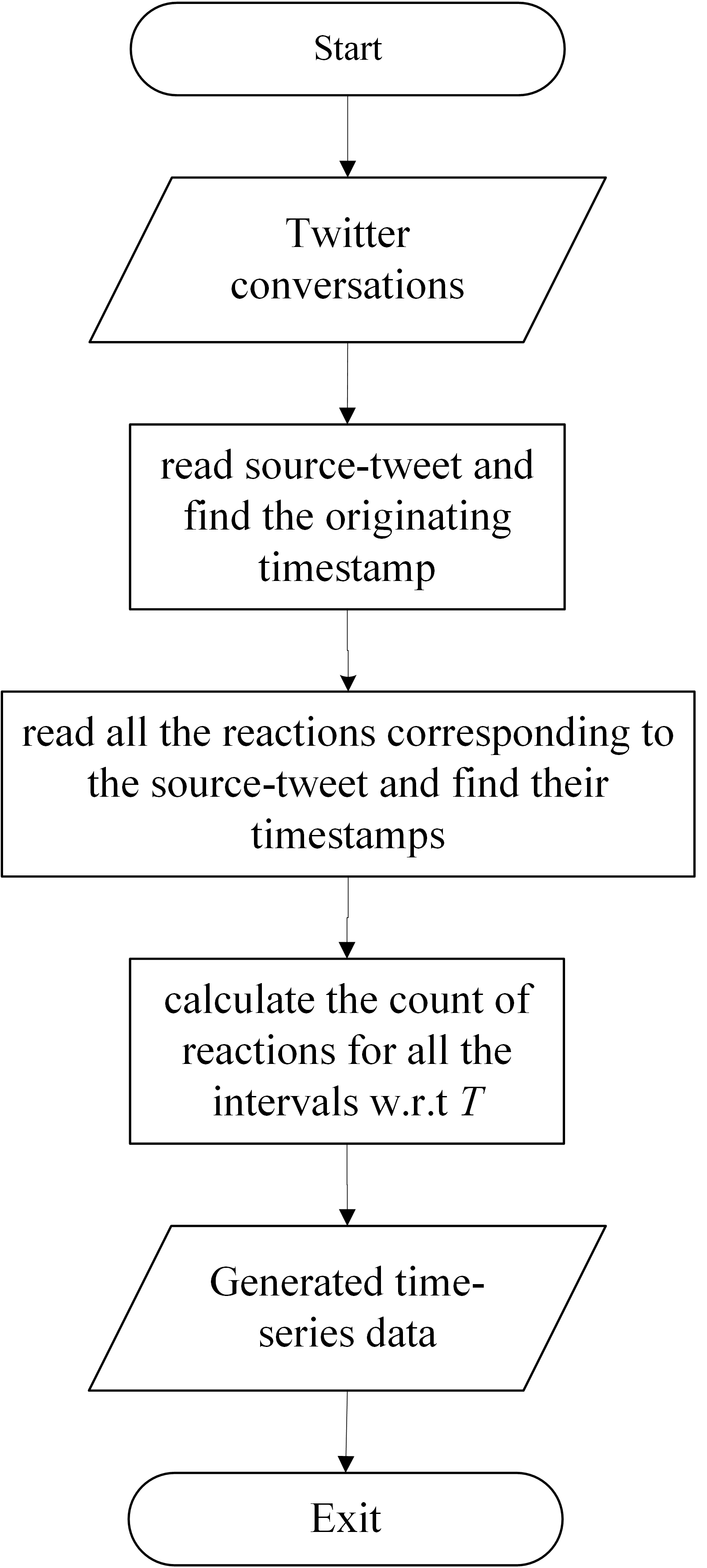}
\caption{The flow chart for transforming Twitter conversations into time-series vectors}
\label{fig:flowchart}
\end{figure}%
The flow chart of transforming Twitter conversations into time-series vectors for all combinations of $E$ and $T$ is given in Fig. \ref{fig:flowchart}.

\subsection{Data pre-processing}
\label{subsec:nntrain}
The second step in our data preparation is to reduce the data sparsity of the time-series data since the vector length of all data samples are decided by the longest conversational sample with respect to $T$. To tackle this problem, we applied sklearn's dimensionality reduction method called \texttt{TruncatedSVD} \cite{scikit-learn}. Next, we normalized the time-series data using sklearn's \texttt{MinMaxScaler} \cite{scikit-learn}, and removed duplicate data samples that are having the same features with different ground truth values.


Finally, we calculated class weights by using sklearn's \texttt{class\_weight} \cite{scikit-learn} library with \textit{balanced} scheme since the PHEME dataset exhibits unbalance class nature. Class weights are used in weighting loss functions during training process, which means higher weight is given to minority class and lower weight to majority class. The class weights are computed using the equation \ref{eq:classwghts} below.
\begin{equation}
class\ weights = \frac{n\_samples}{(n\_classes \times bincount(y))}
\label{eq:classwghts}
\end{equation}
where $y$ represents the actual class labels per sample, $n\_classes$ is the count of unique class label values existing in the dataset, $n\_samples$ is the number of data samples, and $bincount(y)$ counts number of occurrences of each value in $y$ of non-negative integers.

\section{Experimental analysis}
\label{sec:expanalysis}

\subsection{Evaluation metrics}
\label{subsec:metrics}
We used F1-score, which is the weighted average of Precision and Recall scores as the ensemble model's evaluation metric. We considered F1-score metric with micro and macro averaging schemes for evaluating the performances of the ensemble classification models. In general, we calculate F1-score by using equation (\ref{eq:f1}).
\begin{equation}
F1 = 2 \times \frac{precision \times recall}{precision + recall}
\label{eq:f1}
\end{equation}
where precision and recall scores tell the strength of a classifier.

In macro averaging scheme, F1-score is calculated using equation (\ref{eq:macrof1}). Macro F1-score uses precision and recall scores for each class label, and finds their unweighted mean.
In micro averaging scheme, F1-score is determined using equation (\ref{eq:microf1}), and micro F1-score uses global metrics that means precision and recall scores are calculated by counting all the true positives ($TP$), false positives ($FP$), and false negatives ($FN$) across all classes.
\begin{equation}
\begin{aligned}
P_{macro} = \frac{\sum_{i=1}^{n} p_{i}}{n}\\
R_{macro} = \frac{\sum_{i=1}^{n} r_{i}}{n}
\end{aligned}
\label{eq:macropr}
\end{equation}
\begin{equation}
F1_{macro} = 2 \times \frac{P_{macro} \times R_{macro}}{P_{macro} + R_{macro}}
\label{eq:macrof1}
\end{equation}
\begin{equation}
\begin{aligned}
P_{micro} = \frac{\sum_{i=1}^{n} TP_{i}}{\sum_{i=1}^{n} TP_{i} + FP_{i}}\\
R_{micro} = \frac{\sum_{i=1}^{n} TP_{i}}{\sum_{i=1}^{n} TP_{i} + FN_{i}}
\end{aligned}
\label{eq:micropr}
\end{equation}
\begin{equation}
F1_{micro} = 2 \times \frac{P_{micro} \times R_{micro}}{P_{micro} + R_{micro}}
\label{eq:microf1}
\end{equation}
In the above equations, $P$ and $R$ represent precision and recall values for a given averaging scheme (macro or micro), $i$ represents a class label, $p_{i}$ and $r_{i}$ are the precision and recall scores for $i^{th}$ class label. $TP_{i}$, $FP_{i}$, and $FN_{i}$ are the true positives, false positives, and false negatives respectively for  $i^{th}$ class label. $n$ is the total number of classes.

\subsection{Experimental results}
\label{subsec:results}

In TABLE \ref{tab:compare}, we compared our current work's best micro-averaged scores of Precision, Recall, and F1 with our previous works' best micro-averaged results. Clearly, we improved the rumor classification performance by a decent margin with our proposed ensemble based deep learning model in terms of micro-F1. The improvements over Kotteti et al., 2018~\cite{kotteti2018multiple} and Kotteti et al., 2019~\cite{kotteti2019rumor} are $12.5\%$ and $7.9\%$, respectively. The rest of this section discusses  the influence of hyperparameters such as batch input size and learning rate on the classification model's performance.

\begin{table}[htbp]
  \centering
  \caption{Comparison of current work to our previous works}
  \resizebox{\linewidth}{!}{
    \begin{tabular}{c|c|c|c}
    \toprule
    \multirow{2}[4]{*}{Metric} & \multicolumn{2}{c|}{Previous work} & \multicolumn{1}{c}{\multirow{2}[4]{*}{Current work}} \\
\cmidrule{2-3}          & Kotteti et al., 2018~\cite{kotteti2018multiple} & Kotteti et al., 2019~\cite{kotteti2019rumor} &  \\
    \midrule
    Micro-Precision & \textbf{0.949} & 0.564 & 0.643 \\
    \midrule
    Micro-Recall & 0.374 & 0.564 & \textbf{0.643} \\
    \midrule
    Micro-F1 & 0.518 & 0.564 & \textbf{0.643} \\
    \bottomrule
    \end{tabular}}
  \label{tab:compare}
\end{table}

\begin{table*}[htbp]
\begin{minipage}{0.47\textwidth}
  \centering
  \caption{Mean micro averaged F1 testing results of all events that are obtained using leave-one-event-out cross-validation along $T$ by varying learning rate}
    \begin{tabular}{c|c|c|c|c}
    \toprule
    \multirow{2}[4]{*}{\textbf{Time Interval}} & \multirow{2}[4]{*}{\textbf{Learning Rate}} & \multicolumn{3}{c}{\textbf{Micro-F1}} \\
\cmidrule{3-5}          &       & \textbf{I-1} & \textbf{I-2} & \textbf{I-3} \\
    \midrule
    \multirow{3}[6]{*}{2 min} & 5.00E-06 & 0.53986 & 0.52801 & 0.52835 \\
\cmidrule{2-5}          & 1.00E-05 & 0.55231 & 0.53131 & 0.53537 \\
\cmidrule{2-5}          & 1.50E-05 & 0.56656 & 0.53399 & 0.50439 \\
    \midrule
          &       &       &       &  \\
    \midrule
    \multirow{3}[6]{*}{5 min} & 5.00E-06 & 0.43673 & 0.43128 & 0.3936 \\
\cmidrule{2-5}          & 1.00E-05 & 0.43809 & 0.4199 & 0.39489 \\
\cmidrule{2-5}          & 1.50E-05 & 0.44764 & 0.41844 & 0.40408 \\
    \midrule
          &       &       &       &  \\
    \midrule
    \multirow{3}[6]{*}{10 min} & 5.00E-06 & 0.43347 & 0.45515 & 0.46869 \\
\cmidrule{2-5}          & 1.00E-05 & 0.43594 & 0.4086 & 0.41396 \\
\cmidrule{2-5}          & 1.50E-05 & 0.42631 & 0.40358 & 0.41814 \\
    \midrule
          &       &       &       &  \\
    \midrule
    \multirow{3}[6]{*}{30 min} & 5.00E-06 & 0.55092 & 0.43766 & 0.5524 \\
\cmidrule{2-5}          & 1.00E-05 & 0.54492 & 0.42296 & 0.53995 \\
\cmidrule{2-5}          & 1.50E-05 & 0.53428 & 0.45717 & 0.5394 \\
    \midrule
          &       &       &       &  \\
    \midrule
    \multirow{3}[6]{*}{60 min} & 5.00E-06 & 0.55717 & 0.58966 & 0.6116 \\
\cmidrule{2-5}          & 1.00E-05 & 0.61769 & 0.56448 & 0.62146 \\
\cmidrule{2-5}          & 1.50E-05 & 0.619 & 0.55943 & 0.59565 \\
    \bottomrule
    \end{tabular}%
  \label{tab:fixedBSMicro}%
\end{minipage}
\hfill
\begin{minipage}{0.47\textwidth}
  \centering
  \caption{Mean macro averaged F1 testing results of all events that are obtained using leave-one-event-out cross-validation along $T$ by varying learning rate}
    \begin{tabular}{c|c|c|c|c}
    \toprule
    \multirow{2}[4]{*}{\textbf{Time Interval}} & \multirow{2}[4]{*}{\textbf{Learning Rate}} & \multicolumn{3}{c}{\textbf{Macro-F1}} \\
\cmidrule{3-5}          &       & \textbf{I-1} & \textbf{I-2} & \textbf{I-3} \\
    \midrule
    \multirow{3}[6]{*}{2 min} & 5.00E-06 & 0.44878 & 0.38891 & 0.37119 \\
\cmidrule{2-5}          & 1.00E-05 & 0.46329 & 0.39504 & 0.38406 \\
\cmidrule{2-5}          & 1.50E-05 & 0.49849 & 0.38397 & 0.39108 \\
    \midrule
          &       &       &       &  \\
    \midrule
    \multirow{3}[6]{*}{5 min} & 5.00E-06 & 0.41544 & 0.35804 & 0.29844 \\
\cmidrule{2-5}          & 1.00E-05 & 0.42368 & 0.35859 & 0.3125 \\
\cmidrule{2-5}          & 1.50E-05 & 0.42362 & 0.371 & 0.34597 \\
    \midrule
          &       &       &       &  \\
    \midrule
    \multirow{3}[6]{*}{10 min} & 5.00E-06 & 0.34527 & 0.33345 & 0.33153 \\
\cmidrule{2-5}          & 1.00E-05 & 0.35471 & 0.31419 & 0.31294 \\
\cmidrule{2-5}          & 1.50E-05 & 0.34021 & 0.32128 & 0.32075 \\
    \midrule
          &       & \multicolumn{1}{c}{} & \multicolumn{1}{c}{} &  \\
    \midrule
    \multirow{3}[6]{*}{30 min} & 5.00E-06 & 0.37669 & 0.32084 & 0.34954 \\
\cmidrule{2-5}          & 1.00E-05 & 0.38528 & 0.31908 & 0.36389 \\
\cmidrule{2-5}          & 1.50E-05 & 0.38588 & 0.31528 & 0.38077 \\
    \midrule
          &       &       &       &  \\
    \midrule
    \multirow{3}[6]{*}{60 min} & 5.00E-06 & 0.42367 & 0.39506 & 0.45095 \\
\cmidrule{2-5}          & 1.00E-05 & 0.48757 & 0.39201 & 0.45083 \\
\cmidrule{2-5}          & 1.50E-05 & 0.48163 & 0.38864 & 0.43825 \\
    \bottomrule
    \end{tabular}%
  \label{tab:fixedBSMacro}%
\end{minipage}
\end{table*}%

\subsubsection{Fixed batch input size}
The testing results when the batch input size is fixed are shown in Table \ref{tab:fixedBSMicro} and \ref{tab:fixedBSMacro}. These testing results are the mean micro and macro averaged F1 scores of all events that are obtained using leave-one-event-out cross-validation along $T$ by varying learning rate.

\paragraph{micro scores}

From TABLE \ref{tab:fixedBSMicro}, for $T=2$ and $5$ min, the micro-F1 scores of the ensemble Implementation-1 (I-1) are better than that of the  ensemble Implementation-2 (I-2) and Implementation-3 (I-3) across the chosen learning rates. This is due to the fact that it has more ensemble diversity compared with other ensembles, i.e., the presence of base learners designed using Bi-directional RNNs and a model with hybrid architecture that contains a pair of LSTM and GRU layers. In these time intervals, the best scores for the ensemble I-1 are obtained for learning rate $1.50E-05$. 

When $T=10$ min and $T=30$ min, the micro scores performances are mixed. For instance, the ensemble I-1 outperformed ensembles I-2 and I-3 for learning rates $1.00E-05$ and $1.50E-05$ when $T=10$ min. 
For $T=30$ min, the ensemble I-3 achieved maximum micro-F1 score for learning rates $5.00E-06$ and $1.50E-05$. 

For $T=60$ min, the ensemble I-3 outperformed other ensembles in terms of maximum micro-F1 score for learning rates $5.00E-06$ and $1.00E-05$. It is this time interval where all ensembles obtained their maximum micro-F1 scores across $T$ for all chosen time intervals. The overall best micro-F1 score of $62.1\%$ is achieved by the ensemble I-3 for learning rate $1.00E-05$. In this time interval, ensembles I-1 and I-3 are better than that of the ensemble I-2. Again, this is due to more diversity of ensemble I-1 and the base learners in ensemble I-3 with LSTM have better representational power than GRU in ensemble I-2. 

\paragraph{macro scores}

From TABLE \ref{tab:fixedBSMacro}, for $T=2, 5, 10$ and $30$ min, the macro-F1 scores of the ensemble I-1 are better than that of the ensembles I-2 and I-3 across the chosen learning rates. Again, this is due to the presence of more diversified base learners in ensemble I-1 that helped to surpass other ensembles. It is also noticed that when $T=10$ and $30$ min, the performance of the ensemble I-1 dropped down across the learning rates compared to $T=2$ and $5$ min. This is because for longer time intervals, the lengths of time-series data sequences become shorter thus may overlook small propagation patters presented in the time-series data. 

For $T=60$ min, the ensemble I-1 outperformed others in terms of best macro-F1 score for learning rates $1.00E-05$ and $1.50E-05$. When learning rate is $5.00E-06$ the ensemble I-3 surpassed other ensembles. Moreover, in this time interval, for ensembles I-1 and I-2, the results are almost on par with the results that they achieved when $T=2$ min. In this time interval, the ensemble I-3 achieved its overall best performance across $T$. The overall best macro-F1 score is obtained by the ensemble I-1 when $T=2$ min and learning rate of $1.50E-05$.
\paragraph{general observations}
Furthermore, ensembles I-1, I-2, and I-3 better performed in terms of both micro-F1 and macro-F1 scores when $T=60$ min over other time intervals w.r.t the chosen learning rates. The only exception is that the ensemble I-1 performed well in terms of macro-F1 score when $T=2$ min over other time intervals w.r.t the chosen learning rates. In general, both the results are showing us the fact that the performances of ensembles are better when $T$ is either low ($2$ min) or high ($60$ min). This provides a guidance for us to select time interval based on the requirement. For example, if early detection is important we can pick low time interval value. in case of effective prediction, we can go for higher time interval value.

It is also noted that the $10$ min time interval caused most of the ensemble implementations, particularly, the ensemble I-1 to achieve low performance in both micro and macro scores. This may be due to the propagation patterns extracted using this time interval value do not have necessary variations, such that it is harder for classification. Another interesting observation is that ensemble I-3 performs poorly with $5$ min time interval in both micro and macro scores. In this case, using $5$ min time interval caused high data sparsity, which in turn caused LSTM based ensemble I-3 to perform poorly.

\begin{table*}[htbp]
\begin{minipage}{0.47\textwidth}
  \centering
  \caption{Mean micro averaged F1 testing results of all events that are obtained using leave-one-event-out cross-validation along $T$ by varying batch input size}
    \begin{tabular}{c|c|c|c|c}
    \toprule
\multirow{2}[4]{*}{\textbf{Time Interval}} & \multirow{2}[4]{*}{\textbf{Batch Input Size}} & \multicolumn{3}{c}{\textbf{Micro-F1}} \\
\cmidrule{3-5}          &       & \textbf{I-1} & \textbf{I-2} & \textbf{I-3} \\
    \midrule
    \multirow{3}[6]{*}{2 min} & 16    & 0.51013 & 0.48588 & 0.50378 \\
\cmidrule{2-5}          & 32    & 0.55231 & 0.53131 & 0.53537 \\
\cmidrule{2-5}          & 64    & 0.54089 & 0.51473 & 0.52323 \\
    \midrule
          &       &       &       &  \\
    \midrule
    \multirow{3}[6]{*}{5 min} & 16    & 0.48062 & 0.4534 & 0.42757 \\
\cmidrule{2-5}          & 32    & 0.43809 & 0.4199 & 0.39489 \\
\cmidrule{2-5}          & 64    & 0.44371 & 0.46473 & 0.41045 \\
    \midrule
          &       &       &       &  \\
    \midrule
    \multirow{3}[6]{*}{10 min} & 16    & 0.44006 & 0.43332 & 0.48341 \\
\cmidrule{2-5}          & 32    & 0.43594 & 0.4086 & 0.41396 \\
\cmidrule{2-5}          & 64    & 0.43322 & 0.42206 & 0.42 \\
    \midrule
          &       &       &       &  \\
    \midrule
    \multirow{3}[6]{*}{30 min} & 16    & 0.51484 & 0.4542 & 0.50053 \\
\cmidrule{2-5}          & 32    & 0.54492 & 0.42296 & 0.53995 \\
\cmidrule{2-5}          & 64    & 0.55006 & 0.45109 & 0.54926 \\
    \midrule
          &       &       &       &  \\
    \midrule
    \multirow{3}[6]{*}{60 min} & 16    & 0.5789 & 0.5619 & 0.46469 \\
\cmidrule{2-5}          & 32    & 0.61769 & 0.56448 & 0.62146 \\
\cmidrule{2-5}          & 64    & 0.57902 & 0.58571 & 0.64331 \\
    \bottomrule
    \end{tabular}%
  \label{tab:fixedLRMicro}%
\end{minipage}
\hfill
\begin{minipage}{0.47\textwidth}
  \centering
  \caption{Mean macro averaged F1 testing results of all events that are obtained using leave-one-event-out cross-validation along $T$ by varying batch input size}
    \begin{tabular}{c|c|c|c|c}
	\toprule
\multirow{2}[4]{*}{\textbf{Time Interval}} & \multirow{2}[4]{*}{\textbf{Batch Input Size}} & \multicolumn{3}{c}{\textbf{Macro-F1}} \\
\cmidrule{3-5}          &       & \textbf{I-1} & \textbf{I-2} & \textbf{I-3} \\
    \midrule
    \multirow{3}[6]{*}{2 min} & 16    & 0.44335 & 0.39693 & 0.38611 \\
\cmidrule{2-5}          & 32    & 0.46329 & 0.39504 & 0.38406 \\
\cmidrule{2-5}          & 64    & 0.43664 & 0.3692 & 0.36674 \\
    \midrule
          &       &       &       &  \\
    \midrule
    \multirow{3}[6]{*}{5 min} & 16    & 0.4367 & 0.37274 & 0.35665 \\
\cmidrule{2-5}          & 32    & 0.42368 & 0.35859 & 0.3125 \\
\cmidrule{2-5}          & 64    & 0.38173 & 0.32  & 0.32925 \\
    \midrule
          &       &       &       &  \\
    \midrule
    \multirow{3}[6]{*}{10 min} & 16    & 0.34352 & 0.32805 & 0.35568 \\
\cmidrule{2-5}          & 32    & 0.35471 & 0.31419 & 0.31294 \\
\cmidrule{2-5}          & 64    & 0.35004 & 0.31881 & 0.32688 \\
    \midrule
          &       &       &       &  \\
    \midrule
    \multirow{3}[6]{*}{30 min} & 16    & 0.37424 & 0.30853 & 0.3749 \\
\cmidrule{2-5}          & 32    & 0.38528 & 0.31908 & 0.36389 \\
\cmidrule{2-5}          & 64    & 0.38205 & 0.30122 & 0.34899 \\
    \midrule
          &       &       &       &  \\
    \midrule
    \multirow{3}[6]{*}{60 min} & 16    & 0.4266 & 0.37702 & 0.3486 \\
\cmidrule{2-5}          & 32    & 0.48757 & 0.39201 & 0.45083 \\
\cmidrule{2-5}          & 64    & 0.39252 & 0.38015 & 0.47661 \\
    \bottomrule
    \end{tabular}%
  \label{tab:fixedLRMacro}%
  \end{minipage}
\end{table*}%

\subsubsection{Fixed learning rate}
In case of fixed learning rate, the testing results are shown in TABLES \ref{tab:fixedLRMicro} and \ref{tab:fixedLRMacro}. These testing results are the mean micro and macro averaged F1 scores of all events that are obtained using leave-one-event-out cross-validation along $T$ by varying batch input size.

\paragraph{micro scores}
From TABLE \ref{tab:fixedLRMicro}, for $T=2$ and $30$ min, the micro-F1 scores of ensembles I-1 and I-3 are very similar and better than that of the ensemble I-2. This is due to the presence of LSTM layers in both ensembles I-1 and I-3, where in ensemble I-2, there is no base learner with a LSTM layer. In these intervals, w.r.t the chosen batch input sizes, ensemble I-1 achieved the best performance.

When $T=5$ min, the ensemble I-1 outperformed other ensembles for batch input sizes $16$ and $32$. In this time interval, the ensemble I-2 performed better than that of other ensembles for batch size $64$. It is this time interval, where the ensemble I-3 achieved its least micro-F1 scores across all the batch input sizes and $T$, which is the same when batch input size is fixed under micro-averaging scheme. For $T=10$ min, the ensemble I-1 obtained the best micro-F1 scores for batch input sizes $32$ and $64$. And the ensemble I-3 achieved better micro-F1 score over other ensembles for batch input size $16$. In this time interval, the ensembles I-1 and I-2 obtained their least micro-F1 scores across all the batch input sizes and $T$.

When $T=60$ min, the ensemble I-3 outperformed other ensembles for batch input sizes $32$ and $64$. And the ensemble I-1 performed better for batch input size of $16$. It is this time interval, where all ensembles obtained their maximum micro-F1 scores. The overall best micro-F1 score of $64.3\%$ is achieved by the ensemble I-3 for batch input size of $64$. In this case, higher time interval helped the ensembles to surpass their lower time interval micro-F1 scores for almost all of the combinations of batch input size and $T$. Once again, the results show that LSTM backed ensemble I-3 outplayed other ensembles given the advantages of LSTM such as its good gating mechanism and ability to learn long-term dependencies.

\paragraph{macro scores}
From TABLE \ref{tab:fixedLRMacro}, for $T=2$ min, the ensemble I-1 achieved better macro-F1 scores than that of ensembles I-2 and I-3 across all the batch input sizes. In this time interval, the ensemble I-2 obtained its maximum macro-F1 score. When $T=5$ min, the ensemble I-1 outperformed other ensembles in terms of macro-F1 score. Lower time intervals have longer time-series sequences that can better represent variations in propagation patterns of rumors and non-rumors than for higher time interval values. However, lower time intervals may have more data sparsity.

For $T=10$ and $30$ min, the ensemble I-1 achieved better performance than that of other ensembles for batch input sizes $32$ and $64$. However, its performance is significantly dropped compared to lower time interval values. And the ensemble I-3 obtained better performance for batch input size $16$. The ensemble I-2 became weak when $T=30$ min, and ensembles I-1 and I-3 start to show some improvement in their performances compared to $T=10$ min.

In the time interval $T=60$, the ensemble I-1 better performed over other ensembles for batch input sizes $16$ and $32$. And the ensemble I-3 obtained the best macro-F1 score for batch input size $64$. In this time interval, the ensembles I-1 and I-3 obtained their overall maximum macro-F1 scores (i.e. $48.7\%$ and $47.6\%$ respectively) across $T$. Overall, the ensembles support extreme time intervals such as $T=2$ min and $T=60$ min in order to achieve good performance.

\paragraph{general observations}
In case of micro-F1 score, the ensembles I-1, I-2, and I-3 obtained their best micro-F1 scores for $T=60$ min w.r.t the chosen batch input sizes. The only exception is where the micro-F1 score of the ensemble I-3 is lower than its own micro-F1 scores when $T=2, 10$ and $30$ min when batch input size set to $16$. This means that $T=60$ min is appropriate for effective detection of rumors. In case of macro-F1 score, the best performances of the ensembles I-1, I-2, and I-3 are varied for each batch input size across $T$, which means based on the need we can choose an ensemble model and select appropriate time interval. 

As discussed earlier, we have seen the same behavior for $10$ min time interval, which caused most of the ensemble implementations to perform poorly for both micro and macro averaging schemes. In addition to that, the ensemble I-3 again showed low performance in $5$ min time interval under both averaging schemes.

By observing the above results, varying the hyperparameters batch input size and learning rate resulted in producing similar kind of behavior of the ensembles. In general, when micro-averaging is used, both hyperparameter variations supported higher time interval value for better performance. In case of macro-averaging scheme is employed, time intervals $2$ and $60$ min helped ensembles I-1 and I-2 to perform strong. However, ensemble I-3 still achieved better performance when $T=60$ min. As all ensembles are performing good with $60$ min time interval, it is a good choice to achieve decent performance regardless of variations in chosen batch input sizes and learning rates. For $T=60$, the generated time-series data will have lesser data sparsity than that of other values of $T$ that makes the feature space short for the conversation samples. This may be the reason for all ensembles to perform better at higher time intervals. Especially, ensembles with base learners designed using LSTM layers.

Another key observation is that, for all ensembles, $2$ and $60$ min time intervals are shown good performance. However, there is no sweet spot for the ensembles for other values of $T$. This observation is critical in applying the proposed model depending on the goal. For instance, if early detection is needed we can pick small time interval value such as $T=2$ min by sacrificing a little amount of prediction performance. In case of effective prediction is important, we can set time interval to a higher value, for example, $T=60$ min.

\subsection{Discussions}
\label{subsec:summary}
As the PHEME dataset exhibits non-rumor chauvinism (i.e. the dataset contains non-rumor samples almost double the number of rumor samples). Adding more rumor samples to the dataset will help in improving its class balance, and may help classification models to perform better classification. When compared to \cite{kotteti2019rumor}, we noticed that increase in maximum micro and macro averaged F1 scores with addition of two extra events (Gurlitt and Putin Missing events) to the dataset. In case of fixed batch input size, improvement is $5.7\%$ and $0.4\%$ for micro and macro averaging schemes respectively. When learning rate is constant, the improvement is $7.9\%$ for micro averaging scheme. However, maximum macro F1 score is dropped by $0.7\%$. Moreover, even though events Gurlitt and Putin Missing are included in the seven events PHEME dataset only Putin Missing event contributed in adding slightly a greater number of rumor samples to the dataset than Gurlitt event, which is also a supporter of non-rumor group.

In addition to this, our data pre-processing method combined with the proposed model helped in improving our previous best score in \cite{kotteti2019rumor} and achieved $64.3\%$ micro F1 score, which is almost $8\%$ improvement. The performance improvement may seem small, but it is non-trivial to gain huge performances using this dataset, for instance, in \cite{zubiaga2016learning}, extensive feature engineering was conducted for rumor detection problem on social media using the PHEME dataset with five events. The authors focused on extracting complex features such as content-based and social features, and their best F1 scores are $0.606$ and $0.339$ for content-based and social features respectively, and when both feature sets are jointly used the F1 score reached up to $0.607$, which is $0.1\%$ improvement. Again, extensive feature engineering needs long time to be completed as some of the features may not be readily available, having complex feature sets challenge hardware resources, which also increases computational complexity that directly impacts training times of classification models. Nevertheless, given the condition that information spreads rapidly on social media, time-taking labor-intensive feature engineering may not be appropriate.

\section{Related work}
\label{sec:relwork}
Rumor detection on social media is an existing problem in the literature. Many researchers have experimented to find a good solution to this problem. Some of them are, in \cite{liu2016detecting}, authors have explored user-specific features along with content characteristics of social media messages and proposed an information propagation model based on heterogeneous user representation to observe distinctions in the propagation patterns of rumors and credible messages and using it to differentiate them, and their study identifies that rumors are more likely to spread among certain user groups. To predict a document in a social media stream to be a future rumor and stop its spread Qin et al. \cite{qin2018predicting} used content-based features along with novelty-based features and pseudo feedback. In \cite{wang2019rumor}, a sentiment dictionary and a dynamic time series algorithm based Gated Recurrent Unit model is proposed, that identifies fine-grained human emotional expressions of microblog events and the time distribution of social events to detect rumor events.

By treating microblog users' behaviors as hidden clues to detect possible rumormongers or rumor posts Liang et al. \cite{liang2015rumor} proposed a user behavior-based rumor identification schemes, which focuses on applying traditional user behavior-based features as well as authors’ proposed new features that are extracted from users' behaviors to rumor identification task and concluded that rumor detection based on mass behaviors is better than detection based on microblogs'  inherent features. In \cite{kwon2013prominent}, temporal, structural, and linguistic features of social media rumors were explored for rumor classification task and using those features together helped in identifying rumors more accurately. Wu et al. \cite{wu2015false} proposed a graph-kernel based hybrid SVM classifier that can capture high-order (message) propagation patterns as well as semantic features, for example, the topics of the original message for automatically detecting false rumors on Sina Weibo.

As discussed above, most of the works focus on medium to heavy weight feature extraction process, which makes them slow in identifying false information on social media since the fast-paced environment of social media allows a very little amount of time to analyze a piece of information before it propagates all over the network. Our proposed data pre-processing method and ensemble model are capable for this challenge because of the nature of our generated time-series data, and simplicity of classification models' architectures that are part of the ensemble model, and feature extraction process is almost near real-time since our features are creation timestamps of Twitter tweets, which can be extracted and processed without any time delay.

\section{Conclusion}
\label{sec:conl}
In this study, we proposed data pre-processing method and ensemble model for fast detection of rumors on social media. The proposed data pre-processing method transforms Twitter conversations into time-series vectors based on  the tweet creation timestamps, which can be extracted and processed without delay. Furthermore, the generated time-series data is of pure numeric type, which reduces feature set complexity and in turn helps in reducing computational complexity of classification models during their training process. The proposed ensemble model contains several classification models with simplistic yet effective architectures designed using deep learning techniques. By combining the proposed data pre-processing method with the ensemble model, better performance of rumor detection has been demonstrated in the experiments using PHEME dataset. For instance, we improved the classification performance by $7.9\%$ in terms of micro F1 score compared to the baselines.

\section*{Acknowledgment}
\label{sec:acknowledgment}
This research work is supported by the U.S. Office of the Under Secretary of Defense for Research and Engineering (OUSD(R\&E)) under agreement number FA8750-15-2-0119. The U.S. Government is authorized to reproduce and distribute reprints for governmental purposes notwithstanding any copyright notation thereon. The views and conclusions contained herein are those of the authors and should not be interpreted as necessarily representing the official policies or endorsements, either expressed or implied, of the Office of the Under Secretary of Defense for Research and Engineering (OUSD(R\&E)) or the U.S. Government.

\bibliographystyle{IEEEtran}

\bibliography{references}

\end{document}